\pdfoutput=1
\documentclass{article}

\usepackage{arxiv}

\usepackage[utf8]{inputenc} % allow utf-8 input
\usepackage[T1]{fontenc}    % use 8-bit T1 fonts
\usepackage{hyperref}       % hyperlinks
\usepackage{url}            % simple URL typesetting
\usepackage{booktabs}       % professional-quality tables
\usepackage{amsfonts}       % blackboard math symbols
\usepackage{nicefrac}       % compact symbols for 1/2, etc.
\usepackage{microtype}      % microtypography
\usepackage{lipsum}
\usepackage{graphicx}
\usepackage[square,numbers]{natbib}
\usepackage{multirow}
\usepackage{float}
\usepackage{tabu}
\usepackage{caption}
\usepackage{amsmath}
\usepackage{algpseudocode}
\usepackage{hyperref}
\usepackage{authblk}

\usepackage{xcolor}
\usepackage{amssymb}
\usepackage{cancel}
\usepackage{booktabs} 

\usepackage[linesnumbered,lined,ruled,commentsnumbered]{algorithm2e}

\title{HiMem: Hierarchical Long-Term Memory for LLM Long-Horizon Agents}
\author[1]{Ningning Zhang}
\author[1]{Xingxing Yang}
\author[1]{Zhizhong Tan}
\author[1]{Weiping Deng}
\author[1]{Wenyong Wang\thanks{corresponding author}
}
\affil[1]{Macau University of Science and Technology}

\begin{document}

\maketitle

\begin{abstract}
Although long-term memory systems have made substantial progress in recent years, they still exhibit clear limitations in adaptability, scalability, and self-evolution under continuous interaction settings. Inspired by cognitive theories, we propose HiMem, a hierarchical long-term memory framework for long-horizon dialogues, designed to support memory construction, retrieval, and dynamic updating during sustained interactions. HiMem constructs cognitively consistent Episode Memory via a Topic-Aware Event--Surprise Dual-Channel Segmentation strategy, and builds Note Memory that captures stable knowledge through a multi-stage information extraction pipeline. These two memory types are semantically linked to form a hierarchical structure that bridges concrete interaction events and abstract knowledge, enabling efficient retrieval without sacrificing information fidelity. HiMem supports both hybrid and best-effort retrieval strategies to balance accuracy and efficiency, and incorporates conflict-aware Memory Reconsolidation to revise and supplement stored knowledge based on retrieval feedback. This design enables continual memory self-evolution over long-term use. Experimental results on long-horizon dialogue benchmarks demonstrate that HiMem consistently outperforms representative baselines in accuracy, consistency, and long-term reasoning, while maintaining favorable efficiency. Overall, HiMem provides a principled and scalable design paradigm for building adaptive and self-evolving LLM-based conversational agents. The code is available at https://github.com/jojopdq/HiMem.
\end{abstract}
\section{Introduction}

Large language models (LLMs) have demonstrated remarkable progress in language understanding and reasoning, enabling the development of LLM-based agents for complex, multi-turn tasks such as personalized assistance, planning, and long-term decision support \cite{a_finmem, a_kgarevion, a_recommendation}. In realistic interactive settings, however, these agents are required to operate over extended time horizons, where relevant information is scattered across long dialogues and multiple sessions. Despite strong short-term reasoning ability, existing LLM agents still struggle to reliably preserve, organize, and utilize information over long time spans. This limitation has emerged as a fundamental bottleneck for building adaptive and consistent long-horizon conversational agents \cite{t_beyond_goldfish_memory, s_from_human_memory_to_ai_memory, s_beyond_single_turn}.

Recent efforts to address this challenge can be broadly categorized into three directions. Retrieval-augmented generation (RAG) systems introduce external memory stores to fetch relevant information on demand, improving factual grounding \cite{s_conversational_agents_with_LTM, w_hipporag, w_raptor, w_graphrag}. Long-context modeling approaches extend the context window to thousands or even millions of tokens, enabling direct reasoning over extended histories \cite{s_rethinking_memory, w_em-llm, w_memorag, w_readagent}. More recently, structured long-term memory systems have been proposed to persistently store and retrieve dialogue information in compressed or structured forms \cite{s_memory_matters, w_nemori}. While these methods significantly improve efficiency and continuity, they still exhibit systematic limitations when applied to long-horizon interactions \cite{t_fragrel,s_memory_mechanism_of_LLM_agents}.

From both empirical observations and cognitive perspectives, we identify three recurring challenges that existing long-term memory systems struggle to address simultaneously. First, \textbf{semantic misalignment} arises when extracted memories are detached from their original dialogue context, leading to errors in resolving temporal references, coreference, and implicit semantics. Second, most systems rely on \textbf{monolithic or insufficiently hierarchical memory structures}, forcing a trade-off between information fidelity and retrieval efficiency. Fine-grained dialogue logs preserve rich context but incur high retrieval costs, whereas aggressively abstracted representations reduce cost at the expense of critical details needed for reasoning and personalization. Third, memory updates are typically \textbf{static or similarity-driven}, lacking principled mechanisms to revise or correct stored knowledge when new information partially overlaps with, extends, or contradicts existing memories. As a result, long-term consistency degrades over sustained interactions.

Inspired by cognitive theories of human memory \cite{t_schema_memory, s_schema, t_cbt, t_reconsolidation}, we argue that effective long-term memory for LLM agents must satisfy three properties: (i) a \emph{hierarchical structure} that bridges concrete interaction events and abstracted knowledge, (ii) a \emph{unified semantic alignment mechanism} that preserves interpretability across memory representations, and (iii) a \emph{conflict-aware update process} that supports continual self-evolution rather than static accumulation. Based on these principles, we propose \textbf{HiMem}, a hierarchical long-term memory framework designed for long-horizon conversational agents.

HiMem organizes memory into two semantically linked layers. \emph{Episode Memory} preserves fine-grained, temporally grounded interaction segments constructed via a Topic-Aware Event--Surprise Dual-Channel Segmentation strategy, which aligns memory boundaries with both topical shifts and cognitively salient discontinuities. \emph{Note Memory} abstracts stable knowledge such as facts, user preferences, and user profiles through a multi-stage information extraction pipeline. These two memory types form a hierarchical transition from concrete events to compact knowledge representations, enabling efficient retrieval without sacrificing information fidelity. During retrieval, HiMem supports both a hybrid retrieval strategy and a best-effort retrieval strategy that descends from abstract knowledge to concrete events only when necessary. Crucially, retrieval failures are treated as learning signals: HiMem performs conflict-aware Memory Reconsolidation to supplement missing knowledge and revise existing memories, enabling continuous self-evolution over long-term use.

We evaluate HiMem on long-horizon dialogue benchmarks and demonstrate that it consistently outperforms representative baselines in accuracy, consistency, and efficiency. Extensive ablation studies further validate the necessity of hierarchical memory organization, semantic alignment, and conflict-aware updating for robust long-term reasoning.

In summary, this paper makes the following contributions:
\begin{itemize}
    \item We propose \textbf{HiMem}, a hierarchical long-term memory framework that integrates episodic and knowledge-oriented memories to support scalable and adaptive long-horizon conversational agents.
    \item We introduce a Topic-Aware Event--Surprise Dual-Channel Segmentation mechanism and a multi-stage information extraction pipeline to construct cognitively consistent and efficient memory representations.
    \item We design a conflict-aware Memory Reconsolidation mechanism that enables long-term memory systems to self-correct and evolve during sustained interactions.
    \item We provide extensive experimental evidence showing that principled hierarchical design and dynamic updating substantially improve long-horizon reasoning performance.
\end{itemize}

\section{Methodology}
HiMem is a modular long-term memory framework built upon a hierarchical architecture that integrates episodic interaction records with abstracted knowledge representations. It is designed to support efficient retrieval, semantic consistency, and continual memory evolution during long-horizon interactions.
\subsection{Overall Framework}
\begin{figure*}[t]
    \centering
    \includegraphics[width=\linewidth]{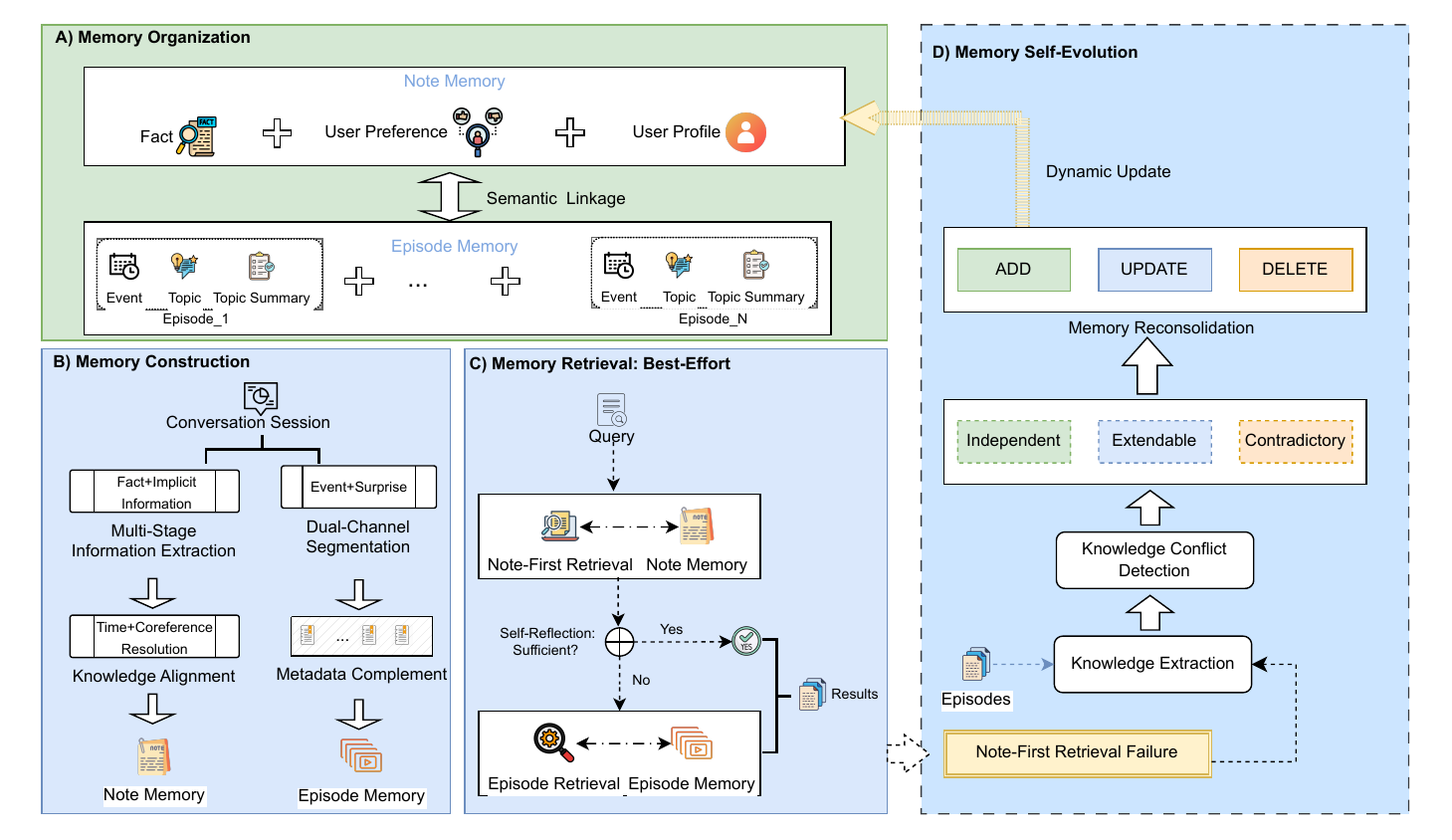}
    \caption{\textbf{Overview of HiMem.} (A) \emph{Memory organization:} a hierarchical connection between Episode Memory and Note Memory. (B) \emph{Memory construction:} pipelines that transform dialogue logs into Episode Memory and Note Memory. (C) \emph{Best-effort retrieval:} hierarchical retrieval in the order of \textbf{Note Memory $\rightarrow$ Episode Memory}, with an LLM assessing evidence sufficiency. (D) \emph{Memory self-evolution:} when evidence from Note Memory is insufficient, the system supplements potentially missing information from Episode Memory and triggers conflict detection and updating.}
    \label{fig:overview}
\end{figure*}
As shown in Figure~\ref{fig:overview}, HiMem consists of three core modules: (i) a hierarchical memory construction module that builds Episode Memory and Note Memory from raw dialogues, (ii) a hierarchical memory retrieval module that supports both hybrid and best-effort retrieval strategies, and (iii) a conflict-aware memory updating module that enables continual self-evolution. Episode Memory preserves fine-grained interaction events, while Note Memory consolidates stable knowledge such as facts, user preferences, and user profiles. The two memory layers are semantically linked to form a hierarchy that bridges concrete experiences and abstract knowledge.
\subsection{Memory Construction}
Memory construction in HiMem follows a multi-stage pipeline that progressively transforms raw dialogue logs into structured long-term memory representations. This pipeline unifies event-level segmentation, knowledge extraction, and semantic alignment to ensure both fidelity and efficiency.
\subsubsection{Episode Memory}
Episode Memory records fine-grained interaction events aligned with topical and cognitive boundaries. Given a dialogue sequence, HiMem segments it into a sequence of non-overlapping episodes. Each episode is represented by a structured record containing an ID, timestamp, topic, topic summary, metadata, and the corresponding dialogue segment.
\paragraph{Dual-Channel Segmentation.}
To obtain cognitively coherent episodes, HiMem adopts a Topic-Aware Event--Surprise Dual-Channel Segmentation strategy. A segmentation boundary is introduced when either (i) a topical shift occurs in discourse goals or subtopics, or (ii) a salient discontinuity is detected, such as an abrupt change in intent or emotional state. These two criteria are fused using an OR rule, producing event units that align with both semantic continuity and cognitive salience.

Segmentation is performed in a single pass, where an LLM jointly evaluates topical and surprise signals and directly outputs the final segmentation. This design yields compact and self-contained episodes that reduce cross-segment interference while preserving critical contextual evidence for downstream reasoning.
\subsubsection{Note Memory}

Note Memory focuses on long-term storage of knowledge-oriented information that remains stable or reusable across interactions. From each dialogue, HiMem extracts three categories of knowledge:
\[
K = \{K_{\text{fact}}, K_{\text{pref}}, K_{\text{profile}}\},
\]
where $K_{\text{fact}}$ denotes objective facts and events, $K_{\text{pref}}$ captures user preferences, and $K_{\text{profile}}$ represents relatively stable user traits.

\paragraph{Multi-Stage Knowledge Extraction.}
Knowledge extraction is decomposed into three stages to avoid semantic collapse. Stage~1 extracts independently interpretable factual and situational units. Stage~2 identifies high-confidence implicit information related to user preferences and profiles without introducing new facts. Stage~3 performs non-destructive normalization, including deduplication, coreference resolution, and temporal normalization, producing aligned knowledge representations suitable for long-term storage. Each aligned knowledge entry is stored as a note, represented as a structured record containing an identifier, the extracted content, a semantic category, and associated metadata.

\subsubsection{Knowledge Alignment}

To maintain semantic consistency across memory layers, HiMem applies a unified alignment process during memory construction. This process includes temporal alignment of relative time expressions, coreference resolution for entity mentions, and extraction of implicit semantic relations. Alignment is selectively applied: Episode Memory prioritizes preserving original dialogue context, while Note Memory emphasizes abstraction and normalization.
\subsection{Memory Retrieval}

HiMem supports two complementary retrieval strategies. In \emph{hybrid retrieval}, the system retrieves information from both Episode Memory and Note Memory to maximize recall. In contrast, \emph{best-effort retrieval} proceeds hierarchically by querying Note Memory first and falling back to Episode Memory only when evidence is deemed insufficient. Retrieved evidence is evaluated by an LLM to assess answerability, and unsupported queries are explicitly marked as unanswerable.

\subsection{Memory Updating and Self-Evolution}

During best-effort retrieval, HiMem employs a fixed LLM-based self-evaluation prompt to assess whether the retrieved evidence is sufficient to answer the query. This evaluation produces a binary judgment (\emph{sufficient} or \emph{insufficient}) under deterministic decoding (temperature = 0) and serves solely as a control signal, without introducing or revising memory content. While related to iterative self-refinement approaches \cite{t_self-refine}, HiMem confines the LLM to deterministic routing and decision control.

Memory reconsolidation is triggered only when two conditions are jointly satisfied: (i) retrieval from Note Memory alone is insufficient, and (ii) the subsequently retrieved Episode Memory provides adequate supporting evidence. This conjunctive trigger grounds updates in episodic context and prevents premature revisions. Although conceptually related to reflective agent frameworks such as \cite{t_reflexion}, HiMem performs structured, evidence-grounded memory operations rather than free-form verbal reflection.

When reconsolidation is triggered, HiMem conducts query-conditioned knowledge extraction over the supporting episodes and compares the extracted information with existing notes. Their relationship is classified as \emph{independent}, \emph{extendable}, or \emph{contradictory}, based on which the system applies ADD, UPDATE, or DELETE operations to revise Note Memory. This typed design avoids indiscriminate overwriting and echoes classic belief revision perspectives \cite{t_knowledge_in_flux}, promoting long-term stability and semantic consistency.

In contrast, episodic memory is treated as immutable: newly constructed episodes are appended chronologically without modification, preserving the temporal integrity of interaction histories.

\subsection{Adaptive Forgetting}

To regulate memory growth under sustained interactions, HiMem optionally employs an adaptive forgetting mechanism based on usage frequency. In this work, forgetting primarily serves as a scalability-oriented mechanism to control memory size and maintain retrieval efficiency, and does not contribute directly to the performance gains reported in our experiments.

\section{Experiments}
\subsection{Datasets}
We evaluate HiMem on \textbf{LoCoMo} \cite{d_locomo}, a benchmark designed to assess long-horizon conversational reasoning under sustained interactions. LoCoMo consists of multi-session dialogues with an average length of approximately 600 turns (around 16K tokens) and spans up to 32 interaction stages, posing significant challenges for long-range dependency modeling and memory management.

The benchmark covers diverse reasoning scenarios, including \emph{Single-Hop} questions answerable within a single session, \emph{Multi-Hop} questions requiring aggregation across distant dialogue turns, \emph{Temporal Reasoning} questions involving implicit or explicit time relations, and \emph{Open-Domain} questions that combine dialogue content with external or commonsense knowledge. Following prior work \cite{w_mem0}, we exclude the Adversarial category from quantitative evaluation, as it focuses on unanswerability detection rather than answer correctness.

\subsection{Evaluation Metrics}

Since different long-term memory systems may apply varying degrees of compression or abstraction over dialogue histories, we adopt a multi-dimensional evaluation protocol to assess answer quality comprehensively. Specifically, following prior work that systematically studies LLM-as-a-Judge and its biases \cite{t_mt-bench, w_secom, w_a-mem}, we use GPT-4o-mini as the LLM judge to compute evaluation score (denoted as GPT-Score) as the primary metric to approximate semantic correctness and consistency, together with F1 to measure lexical overlap.

In addition, for efficiency evaluation, we report \textbf{latency} (Lat.) and \textbf{token consumption} (Tok.). 
Latency is measured as the time required for memory retrieval only, excluding LLM inference and response generation, in order to isolate the efficiency of the memory system.

\subsection{Baselines}

We compare HiMem with representative long-term memory frameworks that cover different design paradigms. \textbf{Mem0} \cite{w_mem0} represents structured memory systems based on atomic factual extraction and graph-based organization. \textbf{SeCom} \cite{w_secom} adopts event-level semantic segmentation and compression to improve contextual completeness. \textbf{A-MEM} \cite{w_a-mem} augments event-level memory with entities, relations, and temporal features to support time-aware retrieval and reasoning. These baselines enable a systematic comparison across retrieval-based, compressed-context, and structured memory approaches.

These baselines are selected based on their compatibility with long-horizon conversational memory, availability of reproducible implementations, and suitability for evaluation under a unified agent interface with comparable inference budgets; methods that primarily target system-level context management or non-dialogue memory access are therefore not included.

\subsection{Settings}

To ensure fair comparison, all methods are evaluated using the same base language model and identical decoding configurations. We use \textbf{GPT-4o-mini} as the backbone LLM and a shared embedding model for vector representations. For each evaluation setting, we conduct three independent trials with fixed prompts.

For baseline-comparative main results, we report mean$\pm$std over multiple runs to reflect run-to-run variability. For auxiliary analyses and ablations (e.g., Table~\ref{tab:knowledge-alignment} and Table~\ref{tab:note-memory}), we report mean values only for compact presentation, as these results are primarily intended to validate relative trends rather than to serve as headline comparisons.

Additional implementation details, including model configurations and hardware specifications, are provided in the Appendix.

\section{Results and Analyses}
\subsection{Main Results}
\begin{table*}[t]
\centering
\setlength{\tabcolsep}{1.5pt}
\renewcommand{\arraystretch}{1.0}
\caption{\textbf{Performance comparison of HiMem and baseline methods on LoCoMo.}
Results are reported as mean (std) over three runs in percentages (\%).
Best results of GPT-Score are shown in \textbf{bold}, and second-best results are underlined.}
\label{tab:main_results}
\begin{tabular}{lcccccccc}
\toprule
\multirow{2}{*}{\textbf{Task}} &
\multicolumn{2}{c}{\textbf{A-MEM}} &
\multicolumn{2}{c}{\textbf{SeCom}} &
\multicolumn{2}{c}{\textbf{Mem0}} &
\multicolumn{2}{c}{\textbf{HiMem}} \\
\cmidrule(lr){2-3} \cmidrule(lr){4-5} \cmidrule(lr){6-7} \cmidrule(lr){8-9}
& GPT-Score & F1 & GPT-Score & F1 & GPT-Score & F1 & GPT-Score & F1 \\
\midrule
Single Hop  & 59.33{\scriptsize (0.51)} & 34.45{\scriptsize (0.46)} & \underline{87.02{\scriptsize (0.35)}} & 23.70{\scriptsize (0.06)} & 75.90{\scriptsize (0.74)} & 53.05{\scriptsize (0.65)} & \textbf{89.22{\scriptsize (0.06)}} & 43.93{\scriptsize (0.24)} \\
Multi Hop   & 40.78{\scriptsize(0.77)} & 20.98{\scriptsize (0.05)} & \underline{59.10{\scriptsize (1.17)}} & 13.21{\scriptsize (0.01)} & 56.62{\scriptsize (2.86)} & 32.90{\scriptsize (1.11)} & \textbf{70.92{\scriptsize (0.77)}} & 28.32{\scriptsize (0.05)} \\
Temporal   & 50.26{\scriptsize (1.55)}  & 35.84{\scriptsize (0.26)} & 33.54{\scriptsize (0.39)} & 4.28{\scriptsize (0.06)} & \underline{68.54{\scriptsize (0.51)}} & 56.37{\scriptsize (0.74)} & \textbf{74.77{\scriptsize (0.25)}} & 22.05{\scriptsize (0.22)} \\
Open Domain & 24.65{\scriptsize (2.14)} & 9.30{\scriptsize (0.50)} & \textbf{60.07{\scriptsize (0.49)}} & 8.57{\scriptsize (0.10)} & 42.36{\scriptsize (0.49)} & 22.70{\scriptsize (0.20)} & \underline{54.86{\scriptsize (1.30)}} & 18.92{\scriptsize (0.45)} \\
Overall     & 51.88{\scriptsize (0.52)} & 30.71{\scriptsize (0.29)} & \underline{69.03{\scriptsize (0.24)}} & 16.77{\scriptsize (0.02)} & 68.74{\scriptsize (0.98)} & 48.16{\scriptsize (0.73)} & \textbf{80.71{\scriptsize (0.21)}} & 34.95{\scriptsize (0.11)} \\
\bottomrule
\end{tabular}
\end{table*}
We evaluate HiMem on the LoCoMo benchmark to assess its ability to preserve, retrieve, and utilize information over long-horizon dialogues. Table~\ref{tab:main_results} reports the performance of HiMem and representative baseline methods across diverse reasoning categories, including Single-Hop, Multi-Hop, Temporal Reasoning, and Open-Domain questions.

Overall, HiMem consistently outperforms all baselines across almost all categories. In particular, HiMem achieves substantial improvements on Multi-Hop and Temporal Reasoning tasks, which require aggregating scattered evidence across long interaction histories. These results indicate that hierarchical memory organization enables more effective modeling of long-range dependencies and semantic consistency than flat or monolithic memory structures. Moreover, the strong performance on Open-Domain questions suggests that HiMem can reliably integrate dialogue-derived knowledge with external or implicit information over extended time spans.

\subsection{Ablation Study: Memory Components}
\begin{table*}[t]
\centering
\setlength{\tabcolsep}{1.5pt}
\caption{\textbf{Ablation study on memory components in HiMem.}
    \textbf{HiMem} includes both \emph{Note Memory} and \emph{Episode Memory};
    \textbf{w/o Note Memory} removes the Note Memory while retaining the Episode Memory;
    \textbf{w/o Episode Memory} removes the Episode Memory while retaining the Note Memory.}
    \label{tab:memory-component}
\begin{tabular}{lccccccc}
\toprule
\multirow{2}{*}{\textbf{Task}} &
\multicolumn{2}{c}{\textbf{HiMem}} &
\multicolumn{2}{c}{\textbf{- w/o Episode}} &
\multicolumn{2}{c}{\textbf{- w/o Note }} \\
\cmidrule(lr){2-3} \cmidrule(lr){4-5} \cmidrule(lr){6-7}
& GPT-Score & F1 & GPT-Score & F1 & GPT-Score & F1 \\
\midrule
Single Hop  & \textbf{89.22{\scriptsize (0.06)}} & 43.93{\scriptsize (0.24)} & 76.50{\scriptsize (0.24)} & 41.09{\scriptsize (0.09)} & \underline{89.02{\scriptsize (0.20)}} & 45.14{\scriptsize (0.03)} \\
Multi Hop   & \textbf{70.92{\scriptsize (0.77)}} & 28.32{\scriptsize (0.05)} & 56.26{\scriptsize (0.33)} & 26.29{\scriptsize (0.12)} & \underline{70.33{\scriptsize (0.44)}} & 26.13{\scriptsize (0.25)} \\
Temporal   & \textbf{74.77{\scriptsize (0.25)}} & 22.05{\scriptsize (0.22)} & 68.12{\scriptsize (0.59)} & 23.65{\scriptsize (0.31)} & \underline{72.48{\scriptsize (0.39)}} & 29.35{\scriptsize (0.16)} \\
Open Domain & \textbf{54.86{\scriptsize (1.30)}} & 18.92{\scriptsize (0.45)} & 48.26{\scriptsize (0.49)} & 22.58{\scriptsize (0.55)} & \underline{48.61{\scriptsize (0.98)}} & 16.81{\scriptsize (0.15)} \\
Overall    & \textbf{80.71{\scriptsize (0.21)}} & 34.95{\scriptsize (0.11)}  & 69.29{\scriptsize (0.05)} & 33.59{\scriptsize (0.06)} & \underline{79.63{\scriptsize (0.22)}} & 36.60{\scriptsize (0.02)}\\
\bottomrule
\end{tabular}
\end{table*}

To examine the contribution of different memory components, we conduct an ablation study by selectively removing Episode Memory or Note Memory from HiMem. The results are shown in Table~\ref{tab:memory-component}.

Removing Episode Memory leads to a pronounced performance degradation across most categories, particularly on Multi-Hop and Temporal Reasoning tasks. This observation highlights the importance of preserving fine-grained contextual evidence aligned with the original interaction process. Without Episode Memory, the system struggles to recover detailed event-level information necessary for tracing complex reasoning chains over long dialogues.

In contrast, removing Note Memory results in a smaller but still consistent performance drop. This suggests that structured knowledge representations primarily serve to accelerate information localization and stabilize semantic anchors, while detailed contextual evidence remains indispensable for coverage and reasoning. Together, these findings demonstrate that Episode Memory and Note Memory play asymmetric yet complementary roles: effective long-term memory systems must balance information fidelity and abstraction rather than relying solely on either raw dialogue context or aggressively compressed representations.
\subsection{Ablation Study: Knowledge Alignment}
\begin{table*}[t]
\centering
\setlength{\tabcolsep}{1.5pt}
\caption{\textbf{Ablation study of the Knowledge Alignment module.} 
For clarity, we denote Knowledge Alignment as KA.
(1) \textbf{HiMem}: alignment applied only to Note Memory;
(2) \textbf{HiMem w/o KA}: alignment disabled for both Note Memory and Episode Memory;
(3) \textbf{Note Memory}: alignment applied to extracted knowledge;
(4) \textbf{Note Memory w/o KA}: extracted knowledge retained without alignment;
(5) \textbf{Episode Memory}: alignment applied to segmented episodes;
(6) \textbf{Episode Memory w/o KA}: segmented episodes retained without alignment.}

    \label{tab:knowledge-alignment}
\begin{tabular}{lccccccccccc}
\toprule
\multirow{2}{*}{\textbf{Method}} & 
\multicolumn{2}{c}{\textbf{Single Hop}} & 
\multicolumn{2}{c}{\textbf{Multi Hop}} &
\multicolumn{2}{c}{\textbf{Temporal}} &
\multicolumn{2}{c}{\textbf{Open Domain}} &
\multicolumn{2}{c}{\textbf{Average}} \\
\cmidrule(lr){2-3} \cmidrule(lr){4-5} \cmidrule(lr){6-7} \cmidrule(lr){8-9} \cmidrule(lr){10-11}
 & GPT-Score & F1 & GPT-Score & F1 & GPT-Score & F1 & GPT-Score & F1 & GPT-Score & F1 \\
\midrule
HiMem & \textbf{89.22} & 43.93 & \textbf{70.92} & 28.32 & \textbf{74.77} & 22.05 & \textbf{54.86} & 18.92 & \textbf{80.71} & 34.95 \\
- w/o KA & 87.51 & 43.75 & 69.86 & 28.53 & 75.18 & 28.14 & 52.08 & 15.96 & 79.50 & 35.98 \\
%Note Memory& \textbf{76.50} & 41.09 & \textbf{56.26} & 26.29 & \textbf{68.12} & 23.65 & \textbf{48.26} & 22.58 & \textbf{69.29} & 33.59 \\
Note Memory& \textbf{66.51} & 35.88 & \textbf{54.26} & 24.92 & \textbf{67.39} & 19.89 & \textbf{50.35} & 23.88 & \textbf{63.44} & 29.79 \\
- w/o KA & 61.79 & 34.16 & 46.81 & 23.57 & 60.12 & 19.77 & 42.71 & 18.52 & 57.51 & 28.25 \\
Episode Memory& 88.31 & 45.53 & 65.25 & 24.76 & 71.55 & 29.08 & \textbf{48.61} & 18.09 & 78.12 & 36.59 \\
- w/o KA & \textbf{89.02} & 45.14 & \textbf{70.33} & 26.13 & \textbf{72.48} & 29.35 & \textbf{48.61} & 16.81 & \textbf{79.63} & 36.60 \\

\bottomrule
\end{tabular}
\end{table*}

We further examine the role of the \textbf{Knowledge Alignment} module by comparing different memory types with and without a unified semantic alignment space. As shown in Table~\ref{tab:knowledge-alignment}, disabling Knowledge Alignment causes a pronounced performance drop for \emph{Note Memory}, indicating that a unified semantic space is crucial for extraction-based memories that do not retain raw dialogue context. Such alignment substantially improves intent understanding and memory localization. In contrast, removing Knowledge Alignment for \emph{Episode Memory} slightly improves performance, suggesting that when segmentation is well-structured, additional semantic fusion may dilute information inherent in raw dialogue. Overall, although extracted knowledge representations are more compact and explicit, they are more sensitive to implicit semantics and coreference resolution, and therefore benefit more from unified semantic alignment.

\subsection{Memory Self-Evolution}

\begin{figure}[t]
    \centering
    \includegraphics[width=\linewidth]{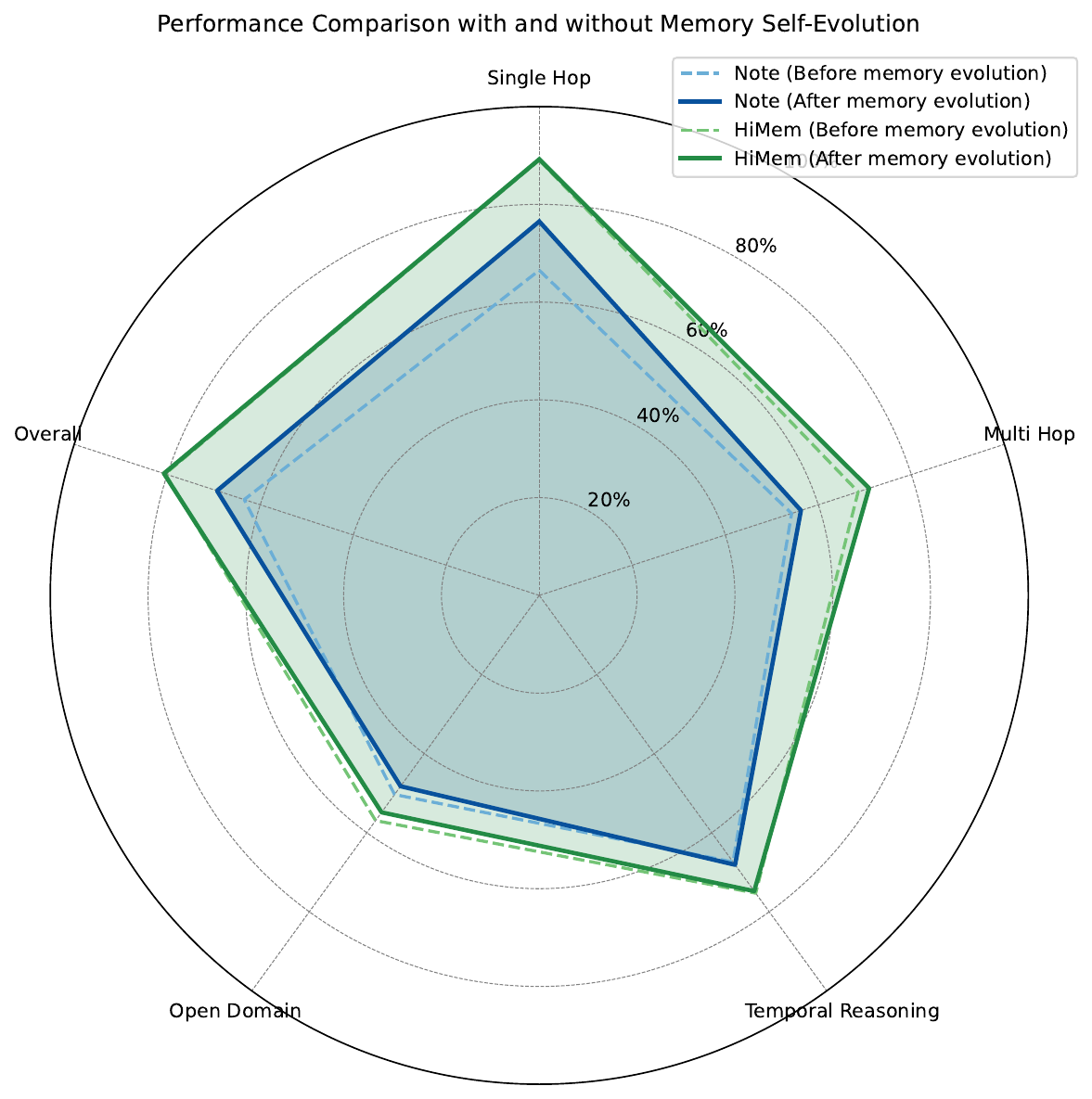}
    \caption{\textbf{Performance comparison (GPT-Score) before and after enabling Memory Self-Evolution 
for Note Memory and HiMem.}
    Memory Self-Evolution is triggered through conflict-aware \emph{Memory Reconsolidation} during best-effort retrieval.}
    \label{fig:self-evolution}
\end{figure}

\begin{table*}[t]
\centering
\setlength{\tabcolsep}{2pt}
\caption{\textbf{Ablation study of the Note Memory.}
For clarity, we denote Knowledge Alignment as KA, and refer to enabling Memory Self-Evolution as ME.}
\label{tab:note-memory}
\begin{tabular}{lcccccccccc}
\toprule
\textbf{Method} &
\multicolumn{2}{c}{\textbf{Single Hop}} &
\multicolumn{2}{c}{\textbf{Multi Hop}} &
\multicolumn{2}{c}{\textbf{Temporal}} &
\multicolumn{2}{c}{\textbf{Open Domain}} &
\multicolumn{2}{c}{\textbf{Average}} \\
\cmidrule(lr){2-3} \cmidrule(lr){4-5} \cmidrule(lr){6-7} \cmidrule(lr){8-9} \cmidrule(lr){10-11}
& GPT-Score & F1 & GPT-Score & F1 & GPT-Score & F1 & GPT-Score & F1 & GPT-Score & F1 \\
\midrule

\textbf{Note Memory} \\
\quad w/o KA
& 61.79 & 34.16 & 46.81 & 23.57 & 60.12 & 19.77 & 42.71 & 18.52 & 57.51 & 28.25 \\
\quad +KA
& 66.51 & 35.88 & 54.26 & 24.92 & 67.39 & 19.89 & 50.35 & 23.88 & 63.44 & 29.79 \\
\quad +KA \& +ME
& \textbf{76.50} & \textbf{41.09} & \textbf{56.26} & \textbf{26.29} & \textbf{68.12} & \textbf{23.65} & \textbf{48.26} & \textbf{22.58} & \textbf{69.29} & \textbf{33.59} \\

\bottomrule
\end{tabular}
\end{table*}

During best-effort retrieval, when \emph{Note Memory} fails to return self-validated results while \emph{Episode Memory} provides sufficient evidence, HiMem triggers the \textbf{Memory Reconsolidation} mechanism. Specifically, the system performs query-conditioned information extraction over retrieved Episode Memory results and supplements missing knowledge in Note Memory through conflict detection and dynamic updating. As shown in Figure~\ref{fig:self-evolution} and Table~\ref{tab:note-memory}, enabling Memory Self-Evolution improves Note Memory performance by approximately \textbf{5.85\%}, which further leads to a slight overall performance gain of about \textbf{0.28\%}. These results demonstrate that Memory Reconsolidation is an effective mechanism for enabling long-term memory self-evolution.

\subsection{Discussion of Extended Analyses}
Additional analyses, including the retrieval strategies, hyperparameter sensitivity, and efficiency trade-offs, are provided in the Appendix.

\section{Discussion}

Long-horizon conversational agents require more than extended context windows or incremental memory accumulation; they critically depend on how information is structured, abstracted, and revised over time. The empirical results consistently demonstrate that hierarchical memory organization is a necessary condition for robust long-term reasoning rather than an optional architectural refinement. Episode Memory and Note Memory play asymmetric yet complementary roles: the former preserves fine-grained contextual evidence aligned with the original interaction process, while the latter consolidates stable, high-frequency knowledge into compact representations that substantially reduce retrieval cost. The performance degradation observed when either component is removed confirms that effective long-term memory systems must balance fidelity and abstraction, instead of relying solely on raw dialogue context or aggressive compression.

Beyond static organization, our findings highlight that memory updating cannot be treated as a purely similarity-driven or append-only process. In long-term interactions, newly observed information often partially overlaps with, extends, or contradicts existing knowledge. HiMem’s conflict-aware Memory Reconsolidation explicitly distinguishes these cases and applies differentiated update strategies, which proves essential for maintaining semantic consistency over time. Importantly, the gains brought by memory self-evolution do not arise from heuristic rewriting, but from a conservative feedback loop between retrieval failure, episodic evidence inspection, and targeted knowledge supplementation.

Finally, the comparison between hybrid and best-effort retrieval strategies indicates that hierarchical memory is not only a representational choice but also an efficiency mechanism. Retrieving abstract knowledge first and descending to concrete events only when necessary achieves a favorable trade-off between accuracy and computational cost, while simultaneously exposing latent information that can drive further memory evolution. Together, these observations suggest that long-horizon LLM agents should treat memory as a dynamic, multi-level system tightly coupled with retrieval and usage, rather than as a static external store.

\section{Conclusion}

This paper proposes \textbf{HiMem}, a hierarchical long-term memory framework for long-horizon dialogues, aimed at addressing several fundamental challenges faced by existing LLM agents in sustained interactions, including memory fragmentation, semantic drift, and the lack of self-evolution capability. Grounded in cognitive theories of human long-term memory, HiMem organically integrates event-level experiences with knowledge-level abstractions, and realizes efficient storage, retrieval, and dynamic updating of long-term information through a structured system design.

Methodologically, HiMem constructs cognitively consistent Episode Memory via \textbf{Topic-Aware Event--Surprise Dual-Channel Segmentation}, providing fine-grained and semantically stable contextual support for complex reasoning tasks. Meanwhile, through a multi-stage information extraction pipeline and selective \textbf{Knowledge Alignment}, high-frequency and stable facts as well as user-specific attributes are consolidated into dense Note Memory representations, significantly reducing retrieval costs while preserving semantic fidelity. Furthermore, HiMem introduces a conflict-aware \textbf{Memory Reconsolidation} mechanism that closes the loop between retrieval and memory updating, enabling continuous correction and evolution of knowledge through usage.

Extensive experiments across multiple long-horizon conversational scenarios systematically validate the effectiveness of these design choices. HiMem consistently outperforms existing methods in terms of accuracy, temporal reasoning, and open-domain understanding. Ablation and analysis studies further reveal that these gains arise from the \emph{synergistic interaction} among hierarchical memory structures, cognitively aligned event segmentation, memory-type-aware semantic alignment, and self-evolution mechanisms, rather than from isolated component-level improvements. In addition, analyses of retrieval modes and hyperparameters demonstrate that HiMem achieves a robust balance between knowledge coverage and system efficiency.

Overall, HiMem’s contributions extend beyond empirical performance improvements. More importantly, it offers a \textbf{practical paradigm for systematically integrating cognitive theories into the design of long-term memory for LLM agents}. By emphasizing memory-type distinctions, structured organization, and usage-driven feedback, this paradigm provides a methodological foundation for building scalable, interpretable, and self-evolving LLM agents. We hope that this work will inspire future research on long-term memory in more complex settings, including multi-agent, multimodal, and richly interactive environments.

\section*{Limitations}

Although HiMem demonstrates stable and significant performance advantages on long-horizon conversational tasks, several limitations remain that warrant further investigation. These limitations do not stem from flaws in the design itself, but rather reflect broader challenges commonly faced by long-term memory systems in realistic interactive settings.

\paragraph{Dependence on LLM Judgment Capabilities.}
First, HiMem relies extensively on the semantic and pragmatic judgment capabilities of the underlying LLM during memory construction and updating, including event segmentation, information extraction, conflict detection, and evidence sufficiency evaluation. While experimental results indicate that such one-shot, rule-constrained judgments are stable and effective in practice, their quality inevitably depends on the capability of the base model. In scenarios involving noisy inputs, metaphorical language, or cross-cultural pragmatic variations, the accuracy of segmentation and knowledge extraction may be affected. Future work could explore incorporating lightweight auxiliary classifiers or uncertainty estimation mechanisms at critical decision points to further enhance robustness under complex linguistic conditions.

\paragraph{Expressive Limits of One-Shot Segmentation.}
Second, HiMem currently adopts a one-shot segmentation strategy, which offers clear advantages in efficiency and controllability, but also imposes an upper bound on expressive capacity. This strategy assumes that the event structure of a conversation can be sufficiently identified through a single global pass. However, in extremely long or highly interleaved dialogues, event boundaries may exhibit hierarchical or recursive structures. Future extensions could investigate multi-granularity or iterative event restructuring strategies, while preserving the simplicity of the current design, to better accommodate non-linear conversational dynamics in Episode Memory.

\paragraph{Conservative Triggers for Knowledge Evolution.}
Regarding memory self-evolution, HiMem primarily relies on retrieval failure or insufficient evidence as triggers for \textbf{Memory Reconsolidation}. While this conservative design promotes stability and avoids unnecessary updates, it may allow certain latent inconsistencies or outdated knowledge to persist if they are not explicitly surfaced during retrieval. Designing more proactive yet noise-resistant evolution triggers remains an open challenge. For example, future work could incorporate user feedback, cross-task consistency checks, or long-term statistical signals to detect and resolve implicit conflicts more effectively.

\paragraph{Limited Evaluation Scope.}
Finally, although HiMem is evaluated on representative long-horizon dialogue benchmarks, the experiments are mainly confined to single-user, text-based interaction scenarios. Real-world long-term interactions often involve multiple users, multimodal inputs, and richer social contexts, which impose additional demands on memory organization and updating. Extending HiMem to multi-agent or multimodal settings, and studying how memories interact, conflict, and propagate across different agents, constitutes an important direction for future research.

Overall, these limitations highlight key research frontiers in advancing long-term memory systems from \emph{usable} to truly \emph{general-purpose}. HiMem provides a viable pathway for systematically integrating cognitive theories into the design of long-term memory for LLM agents. How to further enhance adaptability and generalization while maintaining structural clarity and interpretability remains a central focus for future work.

\section*{Ethical Considerations}
We acknowledge that the development of hierarchical long-term memory systems for LLM agents carries significant ethical responsibilities, particularly concerning data privacy, knowledge integrity, and potential societal impacts. 

\paragraph{Data Privacy and User Profiling}
HiMem is designed to extract and store structured information, including user preferences and profiles, to maintain long-term interaction coherence. In real-world applications, this involves the persistent storage of potentially sensitive personal information. We emphasize that any practical implementation of HiMem should adhere to privacy-by-design principles, such as the General Data Protection Regulation (GDPR). This includes implementing robust data encryption, ensuring transparency regarding what information is being "memorized," and providing users with the "right to be forgotten" by allowing them to inspect and delete specific entries in both Episode and Note Memory. In addition, the datasets used in this study are all publicly available and used in accordance with their respective licenses. 

\paragraph{Knowledge Integrity and Hallucinations}
The "Memory Reconsolidation" mechanism introduces a dynamic self-evolution process where the system updates its internal knowledge based on new interactions. While this improves adaptability, it also poses a risk of "consolidating" hallucinations or incorrect information if the backbone LLM makes erroneous judgments during the conflict-aware update phase. We have mitigated this through a conservative update strategy, but we caution that such systems should not be deployed in high-stakes domains (e.g., medical or legal advice) without human-in-the-loop verification.

\paragraph{Bias Amplification}
As HiMem relies on the semantic understanding and summarization capabilities of pre-trained LLMs, it may inadvertently inherit or amplify biases present in the foundation models during the memory abstraction process (Stage 1-3). We encourage future research to integrate bias-detection filters within the memory extraction pipeline to ensure that the "Notes" stored do not perpetuate harmful stereotypes or unfair social biases.

\paragraph{Intended Use and Transparency}
HiMem aims to foster more meaningful and efficient human-AI collaboration. However, the ability of an agent to form a "long-term bond" through persistent memory could potentially be misused for manipulative purposes. We advocate for full disclosure: users should be explicitly informed when they are interacting with an agent equipped with long-term memory capabilities to manage expectations and ensure informed consent.

\bibliography{references}

\appendix

\section{Related Work}

In recent years, research on long-term memory for LLM agents has progressed along multiple technical directions in parallel, including retrieval-augmented generation, long-context modeling, and structured memory systems. These paradigms respectively emphasize external information access, context capacity expansion, and long-term knowledge organization. However, most existing approaches are developed from isolated design perspectives and lack a unified abstraction to systematically characterize the commonalities, differences, and inherent trade-offs among different memory mechanisms. To this end, we introduce a three-dimensional analytical framework, termed \textbf{Memory Form–Memory Organization–Memory Operation}, which revisits the design space of LLM long-term memory systems from the perspectives of memory unit representation, organizational structure, and dynamic operations, providing a unified basis for comparing different research directions.

\subsection{Memory Form}

\textbf{Memory Form} describes the fundamental representation and granularity of memory units, determining their content structure and serving as the foundational component of long-term memory systems. Early approaches predominantly rely on static segmentation strategies based on dialogue turns or sessions \cite{w_rmm, w_sgmem, w_meminsight}, which often fail to align with semantic boundaries or the temporal evolution of events. Recent work such as SeCom \cite{w_secom} introduces semantic and event boundary detection mechanisms (Event Segmentation) to segment dialogues at the semantic level, leading to notable improvements in semantic coherence and contextual completeness. MemGAS \cite{w_memgas} further provides multiple segmentation modes, allowing systems to select different memory granularities according to task objectives.

Regarding memory content construction, SeCom builds structured memory representations through event segmentation and semantic summarization, while Mem0 constructs fragmented fact units via information extraction. A-MEM \cite{w_a-mem} and THEANINE \cite{w_THEANINE} further enrich memory representations by incorporating multi-dimensional features such as entities, relations, and temporal attributes. Collectively, these approaches explore the balance between information completeness and noise suppression, while supporting temporal reasoning through timestamps or explicit temporal modeling. For instance, A-MEM preserves timestamps to track event order, whereas THEANINE explicitly models temporal dependencies among memory units to capture dynamic semantics.

Despite these advances, existing methods largely focus on explicit semantic segmentation and static feature encoding, without sufficiently modeling implicit semantic dependencies or hierarchical interactions among memory units. In contrast, long-context modeling approaches such as MemGPT \cite{w_memgpt} approach the problem from a system capacity management perspective, emphasizing contextual continuity and memory scheduling rather than semantic structuring of memory content. As a result, they exhibit limitations in fine-grained factual recall and long-term consistency modeling.

Unlike prior work, HiMem refines event granularity through \textbf{Dual-Channel Segmentation} and integrates semantic-level fusion with multi-dimensional feature encoding, unifying explicit structural representations with implicit semantic modeling. This design reflects a shift in memory form from structure-centric storage toward semantic-alignment-centric modeling.

\subsection{Memory Organization}

\textbf{Memory Organization} characterizes how memory units are connected and organized, directly affecting retrieval efficiency and scalability during reasoning. Existing approaches generally follow two main directions. On one hand, structured memory organizations are adopted by methods such as SeCom and A-MEM, which connect memory units linearly along temporal or topical dimensions to maintain semantic continuity. Some frameworks \cite{w_memtree, w_associa, w_mem0, w_g-memory} further introduce tree or graph structures to capture cross-event and cross-topic semantic relations. On the other hand, some approaches draw inspiration from human cognition or operating systems \cite{s_human_inpired_memory,s_cognitive_memory_in_LLMs}. For example, HippoRAG2 \cite{w_hipporag2} combines graph structures with vector spaces to simulate hippocampal indexing mechanisms, enhancing semantic association and retrieval accuracy. MemGPT mimics page caching mechanisms in operating systems, organizing memory hierarchically based on access cost and treating the context window as a high-speed working memory.

However, these methods typically seek compromises within a single organizational structure, making it difficult to simultaneously accommodate diverse task requirements and storage efficiency. Moreover, they primarily focus on data structure design while paying limited attention to explicitly modeling hierarchical differences in memory content. To address this limitation, HiMem adopts a multi-level hybrid organization strategy that constructs a hierarchical long-term memory structure spanning from concrete events to abstract knowledge, based on the information density and abstraction level of memory units. During retrieval, semantic filtering progressively narrows down relevant information, significantly reducing computational cost and noise while preserving high precision.

Beyond structural optimization, HiMem’s key contribution lies in achieving \textbf{Hierarchical Decoupling} in memory organization. By jointly optimizing semantic association, knowledge indexing, and retrieval efficiency within a unified hierarchical framework, HiMem transitions from structure-driven organization to semantic-driven organization.

\subsection{Memory Operation}
\begin{figure*}[t]
    \centering
    \includegraphics[width=\linewidth]{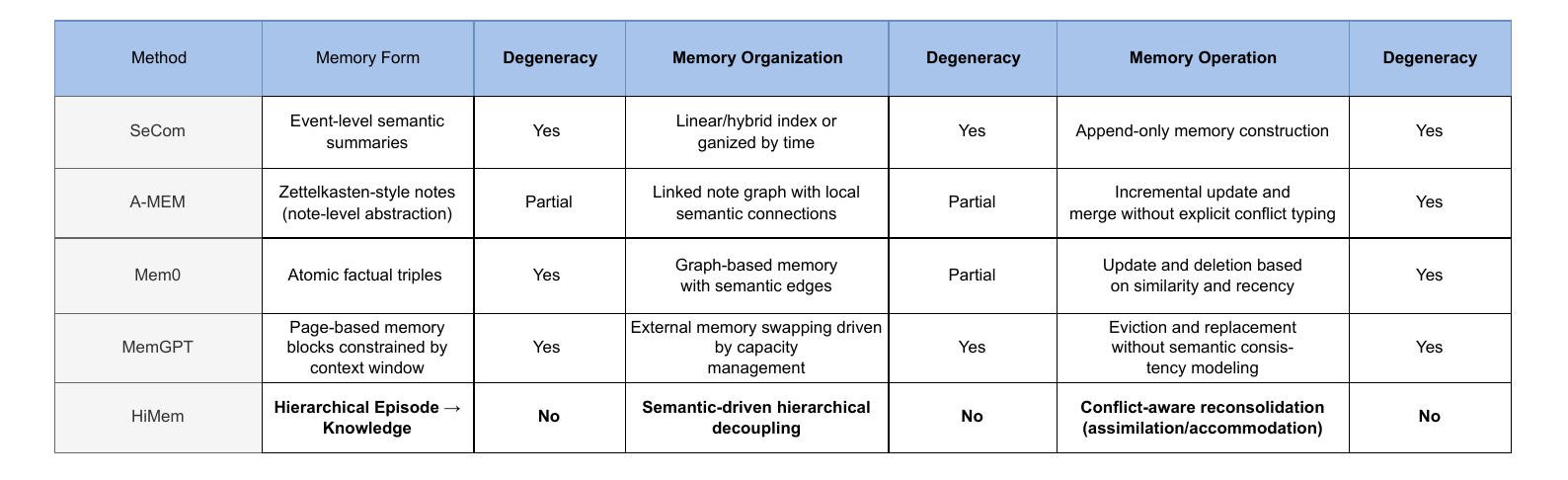}
    \caption{\textbf{Mapping of representative long-term memory systems under the Memory Form–Memory Organization–Memory Operation framework.} The framework characterizes long-term memory systems along three dimensions: memory unit representation, organizational structure, and memory operations. When a dimension collapses into a single fixed design choice that restricts adaptive trade-offs among granularity, structure, or temporal evolution, it is considered to exhibit design degeneration. In contrast, HiMem maintains non-degenerate designs across all three dimensions, enabling more flexible and evolvable long-term memory modeling.}
    \label{fig:memory-3d}
\end{figure*}
\textbf{Memory Operation} focuses on dynamic memory updates and operational mechanisms during usage, which are critical for long-term adaptability and self-evolution. Self-evolving memory is widely regarded as a core capability for long-horizon LLM agents, enabling systems to continuously learn, update, and refine their knowledge structures through sustained interactions \cite{s_LTM_self-evolution}. Several works, including Mem0, A-MEM, and THEANINE, support dynamic updates of memory units to reflect new conversational content. MemoryBank \cite{w_memorybank} introduces a forgetting-curve-based decay mechanism that periodically removes low-frequency or irrelevant information to reduce storage pressure and semantic interference.

MemGPT emphasizes self-management operations of memory: the LLM dynamically reads, writes, and schedules memory across different storage layers based on task requirements, triggering paging and summarization when the context window is constrained, and writing historical information to external storage to maintain contextual continuity. Notably, unlike systems that achieve self-evolution through content-level updates or reconstructions, MemGPT does not directly modify the semantic structure of memory, but instead manages limited context resources via operating-system-style scheduling and compression.

Nevertheless, although these methods support dynamic updates to some extent, they largely remain at the level of content addition or deletion, lacking mechanisms for conflict awareness and semantic reintegration at the knowledge level. When semantic conflicts arise between new and existing information, systems often struggle to decide whether to retain, merge, or revise memories, leading to degraded knowledge consistency or increased computational overhead due to excessive filtering and rewriting.

To address this limitation, HiMem introduces a conflict-aware dynamic evolution mechanism based on memory type differentiation. For memory units that record objective events, whose semantics are relatively stable, conflict detection is unnecessary. In contrast, for knowledge-oriented memories representing user preferences or personal traits, semantic conflicts trigger assimilation or accommodation operations, enabling Memory Reconsolidation and supporting self-correction and continuous evolution. This design closely aligns with cognitive theories of Memory Reconsolidation \cite{t_reconsolidation}, allowing LLM agents to maintain consistency during long-term interactions.

As illustrated in Figure~\ref{fig:memory-3d}, we map representative long-term memory systems into the proposed three-dimensional analytical framework. Most existing approaches avoid design degeneration in only one dimension, while implicitly simplifying the remaining dimensions into fixed design choices. For instance, some methods improve memory form through event segmentation or summarization, but retain static assumptions in memory organization and updating mechanisms. Others enhance organizational flexibility or operational strategies, yet remain constrained in memory form and semantic modeling.

In contrast, HiMem is among the few systems that maintain non-degenerate designs across all three dimensions. It introduces a hierarchical ``event-to-knowledge'' memory form, adopts semantic hierarchy rather than static topology as the organizing principle, and 
explicitly models conflict-aware reconsolidation in memory operations. This three-dimensional non-degeneracy enables HiMem to support more adaptive and evolvable memory management in long-term interaction scenarios.

\section{Implementation Details}
To ensure fair and stable comparisons, all methods are evaluated under identical experimental settings. We use \textbf{GPT-4o-mini} as the base language model for all methods with fixed decoding parameters (\texttt{temperature}=0.0, \texttt{max\_tokens}=8192), set \texttt{top-k}=10 for memory selection during retrieval, and adopt \textbf{all-mpnet-base-v2}~\cite{model_all_mpnet_base_v2} for vector representations. All remaining implementation details strictly follow the official implementations and recommended configurations of the corresponding baselines.

For each benchmark evaluation, we keep model parameters and prompts fixed, run three repeated evaluations, and report the mean and standard deviation of all key metrics across runs. We apply the same evaluation protocol to HiMem and all baselines. In addition, we use GPT-4o-mini as the judge for GPT-Score.

All experiments are conducted on the same hardware environment---a MacBook Pro with Apple M4 Max and 128GB unified memory---to eliminate hardware differences as a confounding factor for efficiency metrics.

For HiMem, we use \textbf{OpenSearch}~\cite{sw_opensearch} as the storage backend for Episode Memory, and adopt \textbf{Qdrant}~\cite{sw_qdrant} to manage knowledge-oriented memory representations in Note Memory.

\section{Extended Analyses}

\subsection{Memory Retrieval Modes}

\begin{table*}[t]
\centering
\setlength{\tabcolsep}{1.5pt}
\caption{\textbf{Performance comparison of HiMem under hybrid and best-effort retrieval strategies.}}
\label{tab:retrieval-performance}
\begin{tabular}{lccccccccccc}
\toprule
\multirow{2}{*}{\textbf{Strategy}} & 
\multicolumn{2}{c}{\textbf{Single Hop}} & 
\multicolumn{2}{c}{\textbf{Multi Hop}} &
\multicolumn{2}{c}{\textbf{Temporal}} &
\multicolumn{2}{c}{\textbf{Open Domain}} &
\multicolumn{2}{c}{\textbf{Average}} \\
\cmidrule(lr){2-3} \cmidrule(lr){4-5} \cmidrule(lr){6-7} \cmidrule(lr){8-9} \cmidrule(lr){10-11}
 & GPT-Score & F1 & GPT-Score & F1 & GPT-Score & F1 & GPT-Score & F1 & GPT-Score & F1 \\
\midrule
hybrid retrieval & \textbf{89.22} & 43.93 & \textbf{70.92} & 28.32 & \textbf{74.77} & 22.05 & \textbf{54.86} & 18.92 & \textbf{80.71} & 34.95 \\
best-effort retrieval & 83.59 & 43.93 & 62.88 & 27.42 & 72.38 & 25.26 & 47.92 & 20.34 & 75.24 & 35.54 \\

\bottomrule
\end{tabular}
\end{table*}

\begin{table*}[t]
\centering
\setlength{\tabcolsep}{3.5pt}
\caption{\textbf{Efficiency comparison of HiMem under hybrid and best-effort retrieval strategies,}
    measured by latency and token consumption.}
\label{tab:retrieval-efficiency}
\begin{tabular}{lccccccccccc}
\toprule
\multirow{2}{*}{\textbf{Strategy}} & 
\multicolumn{2}{c}{\textbf{Single Hop}} & 
\multicolumn{2}{c}{\textbf{Multi Hop}} &
\multicolumn{2}{c}{\textbf{Temporal}} &
\multicolumn{2}{c}{\textbf{Open Domain}} &
\multicolumn{2}{c}{\textbf{Average}} \\
\cmidrule(lr){2-3} \cmidrule(lr){4-5} \cmidrule(lr){6-7} \cmidrule(lr){8-9} \cmidrule(lr){10-11}
 & Lat.(s) & Tok. & Lat.(s) & Tok. & Lat.(s) & Tok. & Lat.(s) & Tok. & Lat.(s) & Tok. \\
\midrule
hybrid retrieval & 1.57 & 1292.53 & 1.55 & 1292.17 & 1.35 & 1177.69 & 1.67 & 1343.25 & 1.53 & 1271.69 \\
best-effort retrieval & 1.63 & 1016.45 & 2.03 & 1257.29 & 1.88 & 1200.50 & 2.66 & 1583.06 & 1.82 & 1134.24 \\

\bottomrule
\end{tabular}
\end{table*}

We compare two retrieval strategies in HiMem: \textbf{hybrid retrieval}, which simultaneously queries Note Memory and Episode Memory, and \textbf{best-effort retrieval}, which first queries Note Memory and descends to Episode Memory only when necessary. As shown in Table~\ref{tab:retrieval-performance}, hybrid retrieval achieves high accuracy while significantly reducing response latency, indicating that the hierarchical memory structure enables parallel localization of relevant information and thus shortens reasoning time. However, as shown in Table~\ref{tab:retrieval-efficiency}, hybrid retrieval incurs the highest token consumption due to aggregating information from both memory layers. In contrast, best-effort retrieval introduces additional latency due to self-evaluation and potential lower-layer queries, but consumes substantially fewer tokens since most queries are resolved at the Note Memory level. These results validate that hierarchical retrieval from abstract knowledge to concrete events achieves an effective trade-off between knowledge coverage and system efficiency.

\subsection{Hyperparameter Analysis}

\begin{figure}[t]
    \centering
    \includegraphics[width=\linewidth]{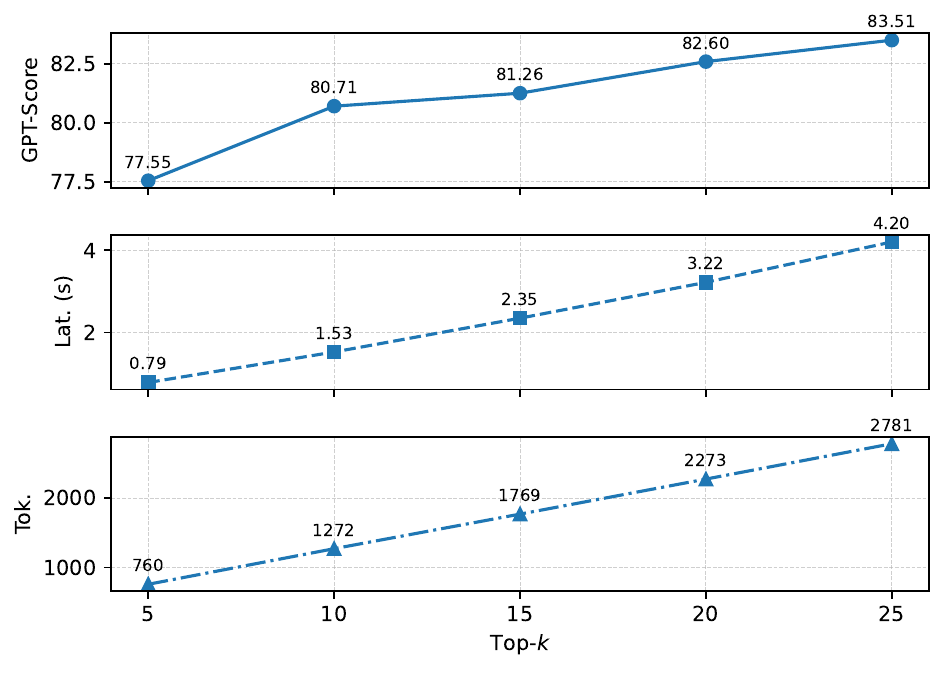}
    \caption{\textbf{Effect of top-k on performance and efficiency.} We report GPT-Score, latency, and token consumption as a function of $k$. Latency reports search latency only.}
    \label{fig:topk}
\end{figure}

We further analyze the effect of the top-k parameter ($k \in \{5,10,15,20,25\}$) on system performance. The results show that increasing $k$ improves retrieval coverage and accuracy, but performance plateaus when $k \geq 10$. Meanwhile, retrieval latency and token consumption increase with larger $k$, reflecting higher retrieval costs. These findings indicate that HiMem, through fine-grained Episode segmentation and multi-stage knowledge extraction, can capture sufficient information within a small retrieval window, achieving optimal performance without expanding the search scope. In contrast, excessively large $k$ introduces irrelevant information and unnecessary processing overhead. This analysis further confirms the efficiency and information-density advantages of HiMem in long-horizon retrieval.
\subsection{Efficiency Analysis}

\begin{table*}[t]
\centering
\setlength{\tabcolsep}{2.5pt}
\caption{\textbf{Efficiency comparison of HiMem and baseline methods on the LoCoMo dataset.}
    The lowest latency and smallest token consumption are highlighted in bold, while the second-best results are underlined. Latency (Lat.) reports search latency only and is measured in seconds (s). Notably, SeCom achieves lower latency by preloading per-sample data into memory prior to inference; therefore, its latency is not directly comparable to that of other methods.}
\label{tab:efficiency_analysis}
\begin{tabular}{lcccccccc}
\toprule
\multirow{2}{*}{\textbf{Task}} &
\multicolumn{2}{c}{\textbf{A-MEM}} &
\multicolumn{2}{c}{\textbf{SeCom}} &
\multicolumn{2}{c}{\textbf{Mem0}} &
\multicolumn{2}{c}{\textbf{HiMem}} \\
\cmidrule(lr){2-3} \cmidrule(lr){4-5} \cmidrule(lr){6-7} \cmidrule(lr){8-9}
& Lat.(s) & Tok. & Lat.(s) & Tok. & Lat.(s) & Tok. & Lat.(s) & Tok. \\
\midrule
Single Hop  & \textbf{0.93{\scriptsize (0.08)}} & 2698.35{\scriptsize (1.20)} & - & 2738.62{\scriptsize (0.05)} & 4.65{\scriptsize (0.13)} & \underline{1586.37{\scriptsize (207.51)}} & \underline{1.57{\scriptsize (0.03)}} & \textbf{1292.53{\scriptsize (0.03)}} \\
Multi Hop   & \textbf{0.91{\scriptsize(0.14)}} & 2715.48{\scriptsize (3.35)} & - & 2742.68{\scriptsize (0.19)} & 3.96{\scriptsize (0.19)} & \underline{1588.96{\scriptsize (209.64)}} & \underline{1.55{\scriptsize (0.03)}} & \textbf{1292.17{\scriptsize (0.15)}} \\
Temporal   & \textbf{0.95{\scriptsize (0.12)}}  & 2697.38{\scriptsize (1.43)} & - & 2612.02{\scriptsize (0.05)} & 4.64{\scriptsize (0.22)} & \underline{1591.95{\scriptsize (209.63)}} & \underline{1.35{\scriptsize (0.02)}} & \textbf{1177.69{\scriptsize (0.11)}} \\
Open Domain & \textbf{0.94{\scriptsize (0.09)}} & 2675.39{\scriptsize (6.14)} & - & 2732.16{\scriptsize (0.73)} & 4.66{\scriptsize (0.19)} & \underline{1498.12{\scriptsize (189.94)}} & \underline{1.67{\scriptsize (0.02)}} & \textbf{1343.25{\scriptsize (0.18)}} \\
Overall     & \textbf{0.93{\scriptsize (0.10)}} & 2699.85{\scriptsize (1.62)} & - & 2712.56{\scriptsize (0.08)} & 4.53{\scriptsize (0.16)} & \underline{1582.51{\scriptsize (207.25)}} & \underline{1.53{\scriptsize (0.03)}} & \textbf{1271.69{\scriptsize (0.05)}} \\
\bottomrule
\end{tabular}
\end{table*}

We evaluate efficiency by comparing response latency and token consumption across methods. Although HiMem introduces semantic fusion and hierarchical retrieval, its latency is not significantly worse than that of other baselines, indicating that the computational overhead of lightweight semantic alignment is minimal. In contrast, by leveraging accurate intent modeling and prior knowledge retrieval, HiMem can more rapidly localize required information. In terms of token consumption, HiMem substantially outperforms all baselines, benefiting from its hierarchical memory structure that prioritizes highly condensed knowledge units and thereby reduces context length and generation burden.

\section{Discussion}

The experimental results systematically demonstrate that HiMem’s advantages in long-horizon conversational tasks do not stem from isolated improvements in individual components. Instead, they arise from a set of \emph{interdependent design decisions} spanning memory representation, organization, and evolution. Together, these decisions form a closed loop from \emph{information acquisition} to \emph{knowledge consolidation} and further to \emph{continuous correction}, enabling stable performance in complex and dynamic conversational environments.

\subsection{Hierarchical Memory as a Fundamental Constraint for Long-Term Dialogue Modeling}

Both the main results and ablation studies consistently indicate that the \textbf{hierarchical memory structure} (Episode Memory + Note Memory) constitutes the core prerequisite for long-term conversational performance. Episode Memory preserves fine-grained contextual information aligned with the original interaction process, allowing the system to accurately trace back critical evidence in tasks such as multi-hop reasoning and temporal dependency modeling. In contrast, Note Memory compresses high-frequency and stable knowledge into dense semantic units through information extraction and structured representation, substantially reducing retrieval cost.

The functional asymmetry between these two memory types explains the patterns observed in the ablation results. Removing Episode Memory leads to a severe performance drop, highlighting that \emph{raw contextual information remains indispensable} for complex reasoning tasks. Removing Note Memory also degrades performance, but to a lesser extent, indicating that structured knowledge primarily serves as a mechanism for accelerating localization and stabilizing semantic anchors. These findings underscore a key insight: \textbf{the effectiveness of long-term memory systems lies not in complete abstraction, but in maintaining a balance between abstraction and fidelity}.

\subsection{The Decisive Role of Dual-Channel Segmentation in Modeling Long-Term Dependencies}

The effectiveness of Episode Memory further depends on how it is constructed. Experimental results show that Episode units built via \textbf{Topic-Aware Event--Surprise Dual-Channel Segmentation} consistently outperform coarse-grained or purely topic-based segmentation strategies in long-term retrieval and reasoning. This suggests that semantic continuity alone is insufficient to capture event boundaries in real conversations; \emph{cognitive-level discontinuities}, such as shifts in emotion, intent, or discourse function, are equally critical signals for long-term memory modeling.

By explicitly incorporating both ``topic continuity'' and ``surprise-driven discontinuity'' criteria at the segmentation stage and fusing them via an OR rule, HiMem generates memory units that better align with human event perception. Such cognitively consistent segmentation reduces cross-segment interference and improves the likelihood that retrieval targets genuinely relevant contexts, which directly manifests as performance gains in Multi-Hop and Temporal Reasoning tasks.

\subsection{Selective Effects of Knowledge Alignment Reveal Memory-Type Differences}

The ablation study on Knowledge Alignment uncovers a more nuanced and informative phenomenon: \textbf{a unified semantic alignment space does not benefit all memory types equally}. For Note Memory, removing Knowledge Alignment results in a substantial performance drop, indicating that extraction-based memories—once detached from raw dialogue context—rely heavily on semantic normalization processes such as coreference resolution and temporal alignment to maintain retrievability and consistency.

In contrast, enabling Knowledge Alignment for Episode Memory may even degrade performance. This observation suggests that when segmentation already achieves sufficient cognitive coherence, further semantic rewriting or fusion of raw dialogue can obscure implicit cues or fine-grained details. This result highlights a central design principle of HiMem: \textbf{semantic alignment should be memory-type aware rather than applied as a uniform preprocessing step}.

\subsection{Memory Reconsolidation as a Key Mechanism for Long-Term Self-Evolution}

Beyond static memory construction, experiments on Memory Self-Evolution further validate the importance of \textbf{Memory Reconsolidation} in sustained interactions. When Note Memory alone cannot support a query but Episode Memory provides sufficient evidence, the system performs targeted information extraction and conflict detection, writing missing knowledge back into Note Memory to correct and enrich its representation.

The effectiveness of this mechanism is evident at two levels. First, Note Memory exhibits a marked performance improvement after self-evolution is enabled, indicating substantive gains in knowledge coverage. Second, the corresponding improvement in overall performance demonstrates that \textbf{retrieval and memory updating should form a feedback loop rather than operate as independent processes}. This stands in contrast to memory systems that only support additive or replacement-based updates, and more closely mirrors human cognition, where memory is reshaped through use.

\subsection{Global Consistency Across Retrieval Modes and Efficiency Analyses}

Finally, comparisons across retrieval modes and the Top-$K$ analysis provide system-level validation of the above design synergies. The complementary behaviors of hybrid and best-effort retrieval in terms of latency and token consumption reflect the flexibility afforded by hierarchical memory under an ``abstraction-first, descend-when-necessary'' strategy. Moreover, the observation that performance saturates at relatively small $K$ values indicates that HiMem’s memory units possess high information density, avoiding inefficient compensation through expanded retrieval scopes.

\subsection{Summary}

In summary, HiMem’s advantages do not arise from larger models or longer context windows, but from a series of \emph{design choices closely aligned with human long-term memory mechanisms}: cognitively consistent event segmentation, memory-type-aware semantic alignment, hierarchically complementary memory representations, and usage-driven Memory Reconsolidation. Together, these elements form a scalable, interpretable, and self-evolving long-term memory framework, providing robust support for sustained interaction and complex reasoning in long-horizon LLM agents.
\end{document}